\documentclass[10pt,twocolumn,letterpaper]{article}

\usepackage{cvpr}              
\usepackage{listings}
\usepackage{enumitem} 
\usepackage{pifont}
\usepackage{hhline}
\usepackage{multirow} 
\usepackage{wasysym}

\newcommand{\tabref}[1]{Tab. ~\ref{#1}}
\newcommand{\figref}[1]{Fig. ~\ref{#1}}
\usepackage{tabularx}

\definecolor{cvprblue}{rgb}{0.21,0.49,0.74}
\usepackage[pagebackref,breaklinks,colorlinks,allcolors=cvprblue]{hyperref}

\def\paperID{11404} 
\def\confName{CVPR}
\def\confYear{2026}

\title{Learning Visual Affordance from Audio}

\author{%
  \textbf{Lidong Lu}\textsuperscript{1}\thanks{Equal contribution}, %
  \textbf{Guo Chen}\textsuperscript{1}\footnotemark[\value{footnote}], %
  \textbf{Zhu Wei}\textsuperscript{2}, %
  \textbf{Yicheng Liu}\textsuperscript{1}, %
  \textbf{Tong Lu}\textsuperscript{1}\thanks{Corresponding author}%
  \\
  \textsuperscript{1}Nanjing University%
  \\
  \textsuperscript{2}China Mobile Communications Company Limited Research Institute%
  \\
  \href{https://jscslld.github.io/AVAGFormer/}{\path{https://jscslld.github.io/AVAGFormer/}}%
}

\begin{document}
\maketitle
\begin{abstract}
We introduce Audio-Visual Affordance Grounding (AV-AG), a new task that segments object interaction regions from action sounds. Unlike existing approaches that rely on textual instructions or demonstration videos, which often limited by ambiguity or occlusion, audio provides real-time, semantically rich, and visually independent cues for affordance grounding, enabling more intuitive understanding of interaction regions. To support this task, we construct the first AV-AG dataset, comprising a large collection of action sounds, object images, and pixel-level affordance annotations. The dataset also includes an unseen subset to evaluate zero-shot generalization. Furthermore, we propose AVAGFormer, a model equipped with a semantic-conditioned cross-modal mixer and a dual-head decoder that effectively fuses audio and visual signals for mask prediction. Experiments show that AVAGFormer achieves state-of-the-art performance on AV-AG, surpassing baselines from related tasks. Comprehensive analyses highlight the distinctions between AV-AG and AVS, the benefits of end-to-end modeling, and the contribution of each component.
\end{abstract}    
\section{Introduction}
\label{sec:intro}
In augmented reality and embodied AI applications, it is crucial not only to identify which object to interact with, but also to determine which specific part of the object to engage with \cite{gibson1977theory}. This makes affordance grounding a key challenge in computer vision, aiming to precisely identify the interactable regions of objects \cite{hassanin2021visual}. Current research largely focuses on achieving affordance grounding through action instructions \cite{luo2022learning,qian2024affordancellm,li2023locate,rai2024strategies,luo2024grounded,chen2024worldafford,jang2024intra} or demonstration videos \cite{nagarajan2019grounded,chen2023affordance,luo2023learning}, but these approaches have limitations, such as imprecise instructions or occlusions in the videos.

So can sounds reveal where to act? In fact, audio can provide a unique supplement in affordance grounding. In daily life, humans can accurately judge interaction regions through auditory cues. For example, the sound of peeling can directly inform us that the handle of a peeler is being used. The advantage of audio lies in its real-time generation, being unaffected by visual interference, and carrying rich semantic information. Objects with similar structures tend to produce similar sounds, which makes audio a more reliable indicator for identifying affordance regions \cite{ren2013auditory,carello2005acoustic,giordano2006material,gibson2014ecological,gould1985designing}. By combining audio and visual modalities, we can significantly enhance the environmental understanding of embodied AI systems, especially in complex or cluttered scenes where visual information may be incomplete or obstructed. This multimodal approach can make AI systems more efficient and intuitive in real-world applications.

\begin{figure}[tp]
    \centering
    \includegraphics[width=\linewidth]{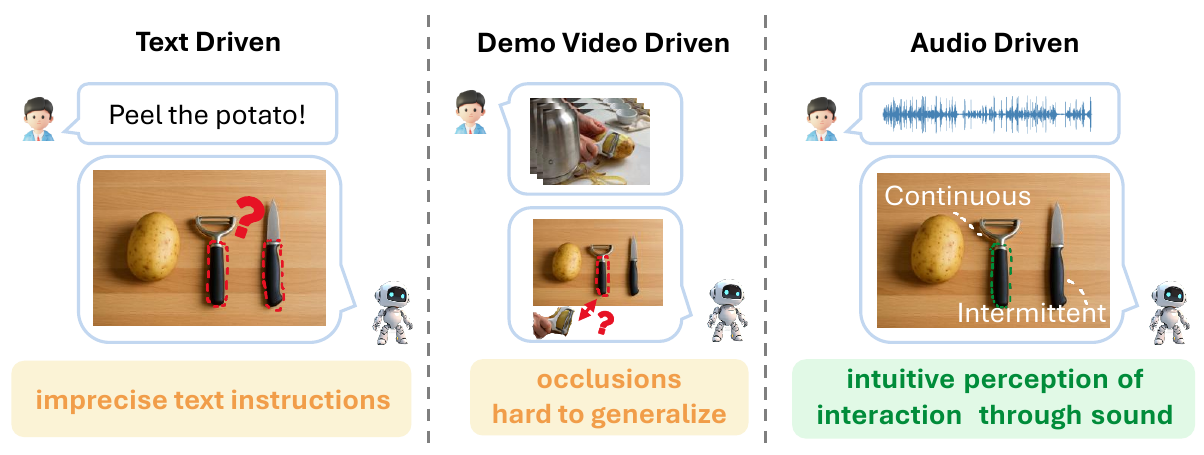}
    \caption{Comparison with text driven~\cite{luo2022learning} and demo video driven~\cite{fang2018demo2vec} affordance grounding, audio driven can help build intuitive perception of interaction regions through sound.}
    \label{fig:0}
\end{figure}

In this paper, we introduce the Audio-Visual Affordance Grounding (AV-AG) task, which aims to segment the parts of an image that are relevant for interaction, based on the action sounds. This task offers both practical value for embodied intelligence and a fine-grained benchmark for multimodal alignment. To support it, we construct AVAGD, a human-annotated dataset containing 12,768 images and 5,203 audio clips covering 97 object and 55 affordance categories with pixel-level segmentation masks. We benchmark both specialized audio-visual segmentation (AVS) models and general multimodal large segmentation models on this dataset, revealing that existing methods perform poorly without retraining, underscoring the unique challenges of AV-AG. We also propose AVAGFormer, an end-to-end baseline featuring a cross-modal mixer for complementary alignment and a dual-head affordance decoder where the function mask guides dependency mask generation for mutual enhancement. Extensive experiments show that AVAGFormer achieves state-of-the-art results on AVAGD. Ablation studies validate each component’s contribution, while results on unseen categories highlight remaining challenges and opportunities for future research.
\section{Related Work}
\label{sec:related}
\begin{table}[t]
\centering
\caption{
\textbf{Comparison with other affordance grounding datasets.} AVAGD offers the richest set of affordance and object categories with well-annotated segmentation mask, and it is the only dataset that provides interaction audio.
}
\resizebox{\columnwidth}{!}{%

\begin{tabular}{ccccccc}
\toprule
Dataset & Year   & \#Obj. & \#Aff. & \#Vis. & \#Aud. &  Annotation \\
\midrule
  UMD\cite{myers2015affordance}    &  2015      &    17     &    7      &   30K    &  - & pixel-level mask \\ 
          CAD120-Aff\cite{sawatzky2017weakly}   &     2017        &     17     &    7      &    3.1K   &  - &  pixel-level mask \\ 
        OPRA\cite{fang2018demo2vec}    &     2018          &     -     &     7     &   20K    &  - &  heatmap \\ 
        EPIC-Hotspot\cite{nagarajan2019grounded}    &     2019          &     31     &     20     &   1.8K    &  -   & heatmap \\ 
        FPHA-Afford\cite{liu2020fpha}    &     2020        &     14     &     10     &   6.6K    &  -  & pixel-level mask \\ 
 AGD20K  \cite{luo2022learning}  &     2022        &     50     &     36     &   24K    & - & heatmap \\ 
        MV-Afford \cite{khalifa2023large}   &     2023        &     37     &     15     &   24K    &  -  & pixel-level mask \\ 
        RAGNet \cite{wu2025ragnet}   &     2025      &  180 &  -   &   273K    &  -  & pixel-level mask \\ 
         AVAGD (Ours)   &     2025        &     97     &     55     &   13K    & 5.2K  & pixel-level mask\\ 
\bottomrule
\end{tabular}
}
\vspace{-0.05in}
\label{tab:tab1}
\end{table}
\noindent \textbf{Affordance Grounding.} Affordance grounding aims to accurately identify the interactable regions of objects in images based on a specific action. Many studies have significantly contributed to this field \cite{luo2022learning,chen2023affordance,luo2023learning,qian2024affordancellm,li2023locate,rai2024strategies,luo2024grounded,chen2024worldafford,jang2024intra}. These methods can be categorized according to action cues. Action-based methods locate affordance regions based on a given action label or instruction. For instance, given the action "press", the model should locate all regions where "pressing" is possible. Notable methods include Cross-View-AG \cite{luo2022learning}, LOCATE \cite{li2023locate}, and AffordanceLLM \cite{qian2024affordancellm}. In contrast, video-based methods locate the affordance region according to the demonstration frames in the video. For example, if a video frame shows someone pressing the first button, the model should predict the first button as the affordance region. Representative methods include Interaction Hotspots \cite{nagarajan2019grounded} and Afformer \cite{chen2023affordance}.

Unlike the previous research, our goal is to enable models to infer affordance regions based on audio cues. In this task, given the sound of pressing a retractable pen, the model should locate the interactable region of the pen for pressing, but exclude other objects that can press, such as a switch, which produce a quite different sound. This presents a greater challenge compared to text-based or video-based methods.

\noindent \textbf{Affordance Grounding Dataset.} As research on affordance grounding has progressed in recent years, several datasets have been proposed \cite{luo2022learning,myers2015affordance,sawatzky2017weakly,fang2018demo2vec,nagarajan2019grounded,liu2020fpha,khalifa2023large,wu2025ragnet}. Early datasets have limitations in terms of data scale and category diversity \cite{myers2015affordance,sawatzky2017weakly,fang2018demo2vec,liu2020fpha}. AGD20K \cite{luo2022learning} significantly enriched the affordance and object categories; however, it only provides sparse heatmap annotations on the test set, which limits the application of supervised learning methods. The dataset proposed by Khalifa et al. \cite{khalifa2023large} has certain advantages in terms of data scale and annotation quality. However, only 35 images contain complex scenes, while the remaining images focus solely on individual objects, creating a domain gap with real-world applications.

Unlike the aforementioned work, our dataset covers diverse domains, including musical instruments, kitchens, daily life, industry and agriculture, sports, and transportation and covers a broader range of affordance and object categories than AGD20K. Most images have  complex backgrounds which better reflect real-world scenarios. Additionally, each image is annotated with pixel-level affordance segmentation mask. Additionally, to our knowledge, this is the first public dataset that provide affordance segmentation mask based on auditory cues. It will promote further research in cross-modal affordance understanding.

\noindent \textbf{Multimodal Query-based Segmentation.} Multimodal query-based segmentation aims to localize and segment image or video regions that correspond to user-specified multimodal queries (e.g., text, audio, or their combinations), and can be viewed as an emerging paradigm for open-vocabulary segmentation empowered by MLLMs~\cite{xu2025qwen2,chen2025eagle,lu2025av,du2025crab}. Early referring image segmentation works~\cite{hu2016segmentation,liu2024grounding} utilize textual expressions as queries and design cross-modal attention mechanisms to guide mask prediction. Recent unified models like  SEEM~\cite{zou2023seem} treat segmentation as a language-conditioned mask generation problem, while Grounded-SAM~\cite{ren2024grounded} combines open-vocabulary prompts with segmentation foundation models~\cite{kirillov2023segment}. Extending beyond text, Audio-Visual Segmentation (AVS)~\cite{zhou2022audio,gao2024avsegformer,li2024selm,yang2023cooperation} explores sound as a natural multimodal query, employing temporal fusion, transformer-based cross-modal decoding, or diffusion modeling to localize sounding sources at the pixel level. Recently, Large Multimodal Segmentation Models such as LISA~\cite{lai2024lisa}, GSVA~\cite{xia2024gsva}, and PSALM~\cite{zhang2025psalm} extend this paradigm by introducing special query tokens that bridge multimodal reasoning and segmentation decoders, enabling flexible integration of diverse input modalities.

\section{Audio-Visual Affordance Grounding}
\subsection{Task Formulation}
The proposed Audio-Visual Affordance Grounding (AV-AG) task can be formalized as an audio-conditioned image segmentation problem. Given an image $\mathcal{I}$ and an audio clip $\mathcal{A}$, the goal of the model $G(\cdot)$ is to identify the functional interaction affordance region and its dependency region in $\mathcal{I}$ that corresponds to $\mathcal{A}$ and generate a binary segmentation mask $\mathcal{M}_{f}$ and $\mathcal{M}_{d}$:
\begin{eqnarray}
\mathcal{G}: (\mathcal{I}, \mathcal{A}) \rightarrow (\mathcal{M}_f,\mathcal{M}_d)
\end{eqnarray}
where $\mathcal{I} \in \mathbb{R}^{H \times W \times 3}$ represents an input RGB image of height $H$ and width $W$. $\mathcal{A} \in \mathbb{R}^{T \times d}$ represents an input audio clip, where $T$ is the number of time steps and $d$ is the audio feature dimension. $\mathcal{M} \in \{0,1\}^{H \times W}$ is the binary segmentation mask, where 
$\mathcal{M}_{(x,y)} = 1$ indicates that pixel $(x,y)$ belongs to the affordance region and $\mathcal{M}_{(x,y)} = 0$ otherwise.

The model $\mathcal{G}(\cdot)$ is trained to learn the mapping between auditory cues and visual affordance regions, ensuring that the predicted segmentation mask $\hat{\mathcal{M}} = \mathcal{G}(\mathcal{I}, \mathcal{A})$ aligns with the ground truth mask $\mathcal{M}^\ast$.

\subsection{Comparison with Other Audio-Visual Tasks}
\begin{table}[h]
\centering
\caption{
\textbf{Comparison with other audio-visual tasks.} In our task, images and audio are asynchronous, and the model is required to perform pixel-level segmentation at the part level.
}

\begin{tabular}{cccc}
\toprule
Task & V-A Rela. & Target & Output   \\
\midrule
  SSL & synchronous & object & heatmap\\
  AVS & synchronous & object & segmentation mask \\
  AV-AG & asynchronous & part & segmentation mask \\
\bottomrule
\end{tabular}


\label{tab:tab5}
\end{table}
As shown in \tabref{tab:tab5}, the most related audio-visual tasks are sound source localization (SSL)~\cite{arandjelovic2017look} and audio-visual segmentation (AVS)~\cite{zhou2022audio,zhou2024audio}. SSL identifies the sound-producing region in an image, typically through coarse heatmaps, while AVS further performs pixel-level segmentation to delineate the entire sounding object. Both tasks rely on temporal synchronization between audio and visual inputs (e.g., a violin sound coinciding with frames of a violin being played), which provides strong contextual cues.

In contrast, our proposed AV-AG task targets finer-grained, part-level affordance grounding rather than object-level localization. It identifies specific functional regions associated with a given audio cue, even when the audio and visual inputs are asynchronous. This asynchrony forces the model to infer potential interactions from static scenes, demanding deeper cross-modal reasoning and understanding of object functionality.

\section{AVAGD Dataset}
\label{sec:dataset}
\begin{figure*}[h]
    \centering
    \includegraphics[width=2\columnwidth]{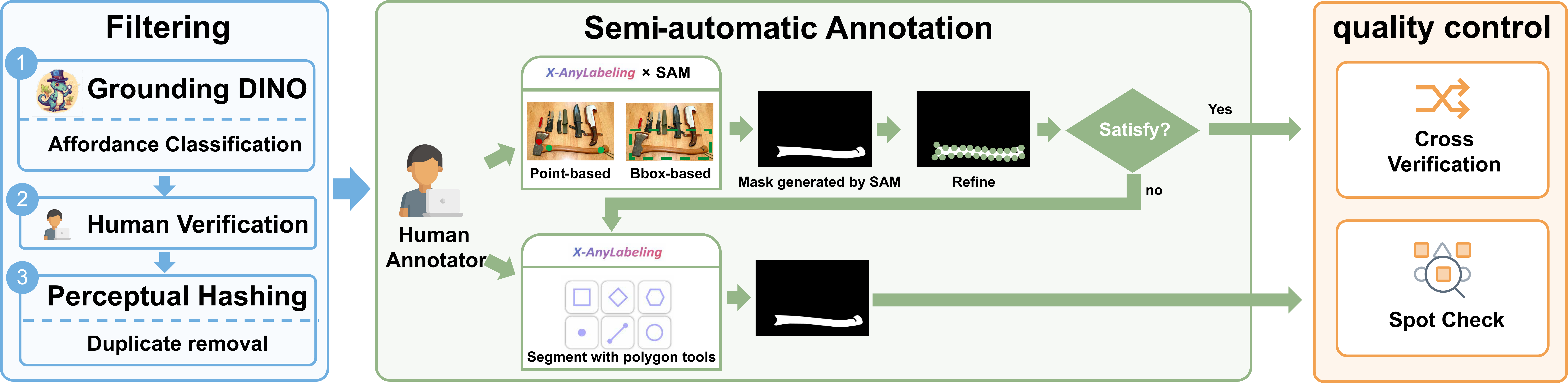}
    \caption{The semi-automatic data annotation pipeline used in the AVAGD dataset}
    \label{fig:annotate}
\end{figure*}

\begin{figure*}[h]
    \centering
    \includegraphics[width=2\columnwidth]{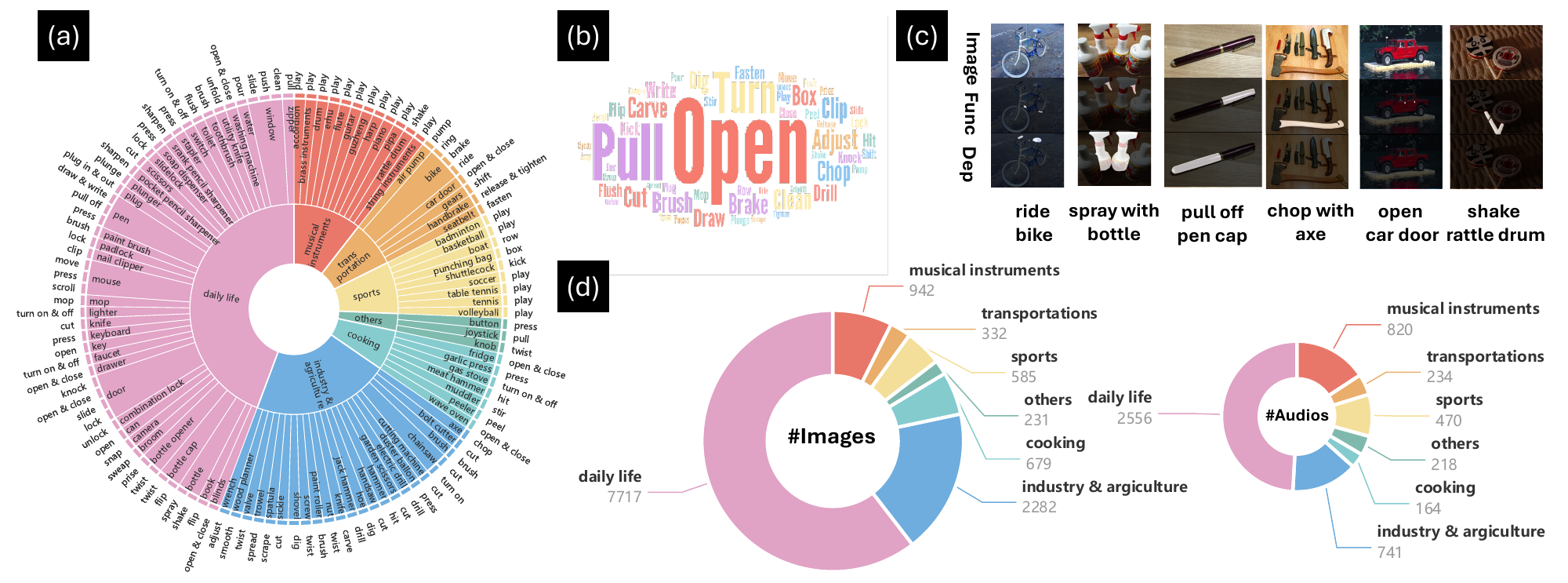}
    \caption{Properties of the AVAGD dataset: (a) Distribution of categories, including 7 domains, 55 affordance categories, and 97 object categories. (b) Word cloud visualization of the affordances in the AVAGD dataset. (c) Annotation samples from the AVAGD dataset. (d) The number of images and audios per object category in the AVAGD dataset.  }
    \label{fig:1}
\end{figure*}

\subsection{Affordance Region Definition}
We categorize affordance regions into two types: function region and dependency region. The function region refers to the area directly involved in achieving an action goal, often undergoing displacement or deformation during interaction, e.g., the cap in the action of pulling off a pen cap. The dependency region provides necessary support or grip for the action, such as the pen body in the same example.

\subsection{Dataset Collection and Annotation}

The Audio-Visual Affordance Grounding Dataset (AVAGD) comprises six domains: musical instruments, cooking, daily life, industry and agriculture, sports, and transportation. To construct the dataset, we primarily source images from ADE20K \cite{zhou2017scene,zhou2019semantic}, Open Images Dataset V7 \cite{kuznetsova2020open}, Products-10K \cite{bai2020products}, and Flickr \cite{flickr}, while additional images from METU-ALET \cite{kurnaz2019alet} are included to mitigate the scarcity of samples in the industry and agriculture domains.

As shown in \figref{fig:annotate}, to efficiently retrieve relevant images from these large-scale datasets, we employ Grounding DINO~\cite{liu2024grounding} for automatic filtering and classification by object category. The selected images are then manually verified to ensure correct classification, assigned affordance labels, and cleaned by removing invalid cases where the interaction region is occluded or in active use. Perceptual Hashing~\cite{krawetz_2011} is further applied to eliminate duplicates across datasets. Finally, human annotators use X-AnyLabeling~\cite{X-AnyLabeling}, assisted by SAM2~\cite{ravi2024sam2}, to generate high-quality affordance segmentation masks according to a pre-defined affordance taxonomy.

For audio, we manually search YouTube using object and affordance categories as keywords. From the retrieved results, we curate 5,203 audio clips by cropping segments with minimal human voices and avoiding overlapping affordance-related sounds.

\subsection{Dataset Statistics}

As shown in \figref{fig:1}, the AVAGD dataset covers 7 domains, 55 affordance categories, and 97 object categories, encompassing common human–object interactions that generate distinctive sounds. Unlike the fixed-length audio in AVSBench, AVAGD features variable-duration audio clips posing a greater challenge for model perception. The dataset includes 12,768 annotated images and 5,203 audio clips, making it the first publicly available affordance grounding dataset with interaction audio. To evaluate zero-shot generalization, the dataset is split into seen and unseen subsets. The unseen subset consists of 29 object-affordance categories for testing, while the seen subset includes 84 categories, where images and audio within each category are further divided into 80\% for training and 20\% for validation.

\subsection{Evaluation Metrics}

Following the AVS setting, we use mean Intersection-over-Union (mIoU) and F-score to evaluate segmentation performance.

\begin{itemize}
    \item \textbf{mIoU}: Measures the pixel-wise overlap between the predicted mask $P$ and the ground truth $Q$:
    \begin{eqnarray}
    \text{IoU}(P, Q) = \frac{|P \cap Q|}{|P \cup Q|}, \quad
    \text{mIoU} = \frac{1}{N} \sum_{i=1}^{N} \text{IoU}(P_i, Q_i)
    \end{eqnarray}
    where $N$ is the number of samples.

    \item \textbf{F-score}: Balances precision and recall, particularly for imbalanced affordance regions:
    \begin{eqnarray}
    \text{F-score} = \frac{1}{N} \sum_{i=1}^{N} 
    \frac{(1 + \beta^2) \cdot \text{Precision}_i \cdot \text{Recall}_i}
    {\beta^2 \cdot \text{Precision}_i + \text{Recall}_i}
    \end{eqnarray}
    with $\beta^2 = 0.3$ to emphasize recall.
\end{itemize}

\subsection{Comparison with Existing Datasets}

\tabref{tab:tab1} presents a comprehensive comparison between AVAGD and existing datasets. AVAGD significantly advances affordance grounding research by providing the largest set of object and affordance categories, along with a substantial number of images covering diverse and complex real-world scenes. All images are human-annotated with high-quality, pixel-level affordance masks. Notably, AVAGD is the first dataset to include auditory cues linked to human–object interactions, enabling multimodal studies on how sound complements visual affordance perception.

In addition, our annotation scheme explicitly distinguishes between functional regions and contextual dependency regions that indirectly support the affordance. This fine-grained labeling enhances interpretability and supports embodied intelligence applications, where recognizing both direct and indirect affordance cues is essential for effective interaction.

\subsection{Evaluation with Existing Methods}
\begin{table}[h]
\centering
\caption{Zero-shot evaluation results of the AVSBench-object S4 pre-trained model on the AV-AG task, demonstrating the uniqueness of our proposed task.}
\resizebox{\columnwidth}{!}{%
\begin{tabular}{c|cc|cc}
\toprule
\multirow{2}{*}{Method} & \multicolumn{2}{c|}{Seen} & \multicolumn{2}{c}{Unseen}  \\
 &   $\text{mIoU}$ $\uparrow$ & $\text{F}$ $\uparrow$&  $\text{mIoU}$ $\uparrow$ & $\text{F}$ $\uparrow$\\
\midrule
CAVP~\cite{chen2024unraveling} & 19.91 & 31.71 & 9.72 & \underline{19.73}\\
Selm~\cite{li2024selm} & \textbf{28.34} & \textbf{37.33} & \underline{11.35} & \textbf{20.99}\\
COMBO-AVS~\cite{yang2023cooperation} & \underline{27.22} & \underline{37.09} & 10.97 & 19.25 \\
AVSegFormer~\cite{gao2024avsegformer} & 25.83 & 30.56 & \textbf{12.12} & 16.32 \\
\bottomrule
\end{tabular}
}
\label{tab:tab6}
\vspace{-3mm}
\end{table}
As shown in \tabref{tab:tab6}, we evaluate several state-of-the-art AVS models on the AV-AG task without additional training or adaptation. Each image is repeated five times, and the corresponding audio is repeated or resampled to 5 seconds following the AVSBench-object S4~\cite{zhou2022audio} protocol. The mIoU and F1 scores are averaged over the five frames.

All AVS models perform poorly on AV-AG, indicating limited generalization from object-level AVS to part-level AV-AG. This gap stems from the fundamental task difference: AVS focuses on sound-emitting objects, whereas AV-AG targets interactable parts that imply potential actions even without sound. The shift from object- to part-level reasoning increases semantic complexity, as it requires understanding functionality rather than acoustic activity. Moreover, existing AVS models lack explicit supervision for affordance or part segmentation, leading to misaligned representations and a notable performance drop on AV-AG.

\begin{table*}[h]
\centering
\caption{Zero-shot evaluation results of the large multimodal segmentation model on the AV-AG task.}
\resizebox{2\columnwidth}{!}{%
\begin{tabular}{ll|cccc|cccc}
\hline
\multirow{2}{*}{Method} & \multirow{2}{*}{Paradigm}  
& \multicolumn{4}{c|}{Seen} 
& \multicolumn{4}{c}{Unseen} \\
& & $\text{mIoU}_{f}\!\uparrow$ 
& $\text{F}_{f}\!\uparrow$ 
& $\text{mIoU}_{d}\!\uparrow$ 
& $\text{F}_{d}\!\uparrow$ 
& $\text{mIoU}_{f}\!\uparrow$ 
& $\text{F}_{f}\!\uparrow$ 
& $\text{mIoU}_{d}\!\uparrow$ 
& $\text{F}_{d}\!\uparrow$ \\
\hline
\multicolumn{9}{l}{\textbf{\textit{Large Multimodal Segmentation Models with Text Queries}}} \\
\hline
PolyFormer-B~\cite{liu2023polyformer} & Boundary Point-based
  & 16.45 & 25.97 & 17.87 & 23.66
  & 17.61 & \textbf{29.35} & 18.13 & 25.35 \\
LISA~\cite{lai2024lisa} & Hidden State-based
  & 8.97 & 18.51 & 7.36 & 13.19
  & 7.59 & 16.44 & 7.05 & 9.44 \\
HiMTok~\cite{wang2025himtok} & Mask Token-based
  & \underline{20.29} & \underline{30.94} & 22.97 & 28.08
  & \textbf{18.33} & 28.21 & \underline{23.36} & \underline{31.07} \\
SAM4MLLM~\cite{chen2024sam4mllm} & Others 
    & 18.47 & 28.51 & \underline{23.36} & \underline{31.82} 
    & \underline{17.35} & \underline{28.98} & 22.93 & 30.76 \\

\hline
\multicolumn{9}{l}{\textbf{\textit{Large Multimodal Segmentation Models with Audio Queries}}} \\
\hline
Crab~\cite{du2025crab}     & Hidden State-based & \textbf{31.54} & \textbf{44.92} & \textbf{42.01} & \textbf{59.33} & 16.72 & 27.91 & \textbf{31.38} & \textbf{44.73} \\
\hline
\end{tabular}
}

\label{tab:tab7}
\end{table*}

We further evaluate several representative large multimodal segmentation models on the AV-AG task, as shown in \tabref{tab:tab7}. Results show that two-stage text-based methods consistently lag behind end-to-end audio-based ones. The performance gap mainly stems from semantic drift and accumulated errors in query generation, as LLM-produced textual descriptions often misalign with the visual or auditory context, weakening cross-modal correspondence during segmentation.

\section{A Strong Baseline: AVAGFormer}
\begin{figure*}[h]
    \centering
    \includegraphics[width=\linewidth]{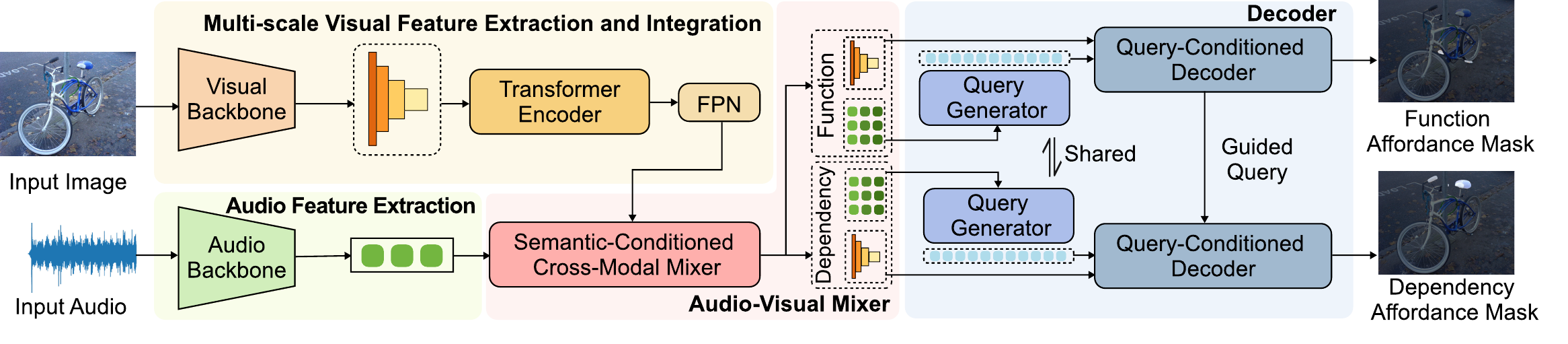}
    \caption{Overview of our proposed AVAGFormer. It consist of three key components: visual and audio feature extraction and integration, audio-visual mixer, and affordance decoder.}
    \label{fig:2}
\end{figure*}

\label{sec:method}

\subsection{Overall Architecture}
As illustrated in \figref{fig:2}, AVAGFormer first extracts audio and visual features using pretrained backbones. Visual features are enhanced by a Transformer Encoder and FPN to fuse global context and multi-scale spatial details. The semantic-conditioned cross-modal mixer then aligns and integrates audio and visual features, producing function- and dependency-specific representations conditioned on audio cues. Finally, a dual-head query-conditioned decoder generates function and dependency affordance masks.
\subsection{The Encoder}
We use pretrained backbones to extract features from both audio clips and images. For the audio clip, we first apply STFT \cite{allen1977unified} to obtain its spectrogram. Then, a pretrained backbone is used to extract audio feature $\mathcal{F}_{\text{audio}} \in \mathbb{R}^{T \times 128}$. For the image, we extract multi-scale features $\mathcal{F}_{\text{visual}_i}\in \mathbb{R}^{H_i \times W_i \times C_i}$, where $(H_i,W_i)=(H,W)/2^{i+1}$, and $i \in \{1,2,3,4\}$.
\subsection{Semantic-Conditioned Cross-Modal Mixer}
\label{sec:method.1}

The semantic-conditioned cross-modal mixer introduces task-specific semantic cues to guide bidirectional fusion between visual and audio modalities. It leverages cross-attention to align and reconstruct modality features, generating function- and dependency-aware representations that facilitate semantic grounding.

Given the input visual features $X^v \in \mathbb{R}^{B \times N_v \times C_v}$ and audio features $X^a \in \mathbb{R}^{B \times N_a \times C_a}$, both are linearly projected into a shared embedding space as $\tilde{X}^v$ and $\tilde{X}^a$ with a unified dimension $C=C_a$. To inject semantic conditioning, we compute spatial and temporal magnitudes $\bar{m}_v, \bar{m}_a \in \mathbb{R}^{B \times 1}$ and concatenate them with semantic prompts $p_t^v, p_t^a \in \mathbb{R}^{C}$ for task type $t \in \{\text{func}, \text{dep}\}$. These combined signals are processed by projection networks to form adaptive attention biases:
\begin{eqnarray}
\begin{aligned}
\text{Bias}^v_t &= f^v(\text{Concat}(p^v_t, \bar{m}_v, \bar{m}_a)), \\
\text{Bias}^a_t &= f^a(\text{Concat}(p^a_t, \bar{m}_v, \bar{m}_a)).
\end{aligned}
\end{eqnarray}

Although $\bar{m}_v$ and $\bar{m}_a$ are global scalars, they modulate the fusion strength according to the overall activation of each modality.

Semantic-conditioned cross-attention is performed in both directions. For audio attending to vision, the attention map is computed as:
\begin{align}
\text{Attn}_{a \leftarrow v} = \text{Softmax}\left( \frac{Q^a (K^v)^T}{\sqrt{d}} + \text{Bias}^a_t \right),
\end{align}

and the fused audio features are obtained by $\hat{X}^a = \text{Attn}_{a \leftarrow v} V^v$. Similarly, vision attends to audio through $\hat{X}^v = \text{Attn}_{v \leftarrow a} V^a$. To control the degree of fusion, task-aware gating functions are defined as $g^v = \sigma(W^v_g p^v_t)$ and $g^a = \sigma(W^a_g p^a_t)$, leading to the final gated outputs:
\begin{eqnarray}
\begin{aligned}
X^v_t &= g^v \odot \hat{X}^v + (1-g^v) \odot \tilde{X}^v, \\
X^a_t &= g^a \odot \hat{X}^a + (1-g^a) \odot \tilde{X}^a,
\end{aligned}
\end{eqnarray}

where $\odot$ denotes element-wise multiplication. The visual output $X^v_t$ is then projected back to its original channel dimension $C_v$ and reshaped into spatial form:
\begin{align}
X^v_t \in \mathbb{R}^{B \times C_v \times H \times W}.
\end{align}

This fusion is independently applied across multiple visual scales and semantic types, producing task-specific visual and audio representations conditioned on semantic prompts that strengthen cross-modal alignment and improve affordance grounding performance.

\subsection{Dual-head Affordance Decoder}
\begin{figure}[h]
    \centering
    \includegraphics[width=\linewidth]{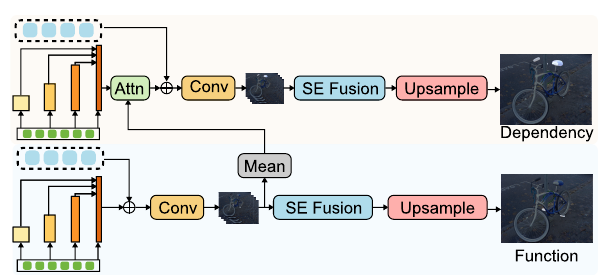}
    \caption{Architecture of the dual-head affordance decoder.}
    \label{fig:head}
\end{figure}
As shown in \figref{fig:head}, multi-scale audio-visual features from the semantic-conditioned cross-modal mixer are first processed by a multi-scale decoding module, which performs cross-modal fusion at each scale. The decoded features are upsampled and aligned to the highest resolution to form base semantic features. Semantic prompts are then transformed into semantic queries that explicitly guide function region mask generation, ensuring semantic consistency.

The predicted function mask subsequently serves as a condition in a mask-conditioned attention module to infer the dependency region, producing its segmentation mask. Finally, segmentation results from all queries are aggregated through a SE module~\cite{hu2018squeeze} and progressively upsampled to the original resolution to yield the final prediction.

\subsection{Objective Function}
We adopt an auxiliary loss~\cite{xu2021leveraging,navon2020auxiliary} to provide additional supervision. The averaged candidate mask from multiple queries is upsampled via bilinear interpolation to generate an auxiliary mask.

For function affordance prediction, where foreground regions are typically small, we combine weighted Dice and Focal losses to form the auxiliary loss $\mathcal{L}_{\text{aux}}$, and employ an IoU loss $\mathcal{L}_{\text{IoU}}$ for the final prediction:
\begin{eqnarray}
\mathcal{L}_{\text{func}} = \mathcal{L}_{\text{IoU}} + \lambda \mathcal{L}_{\text{aux}}.
\end{eqnarray}

Dice loss preserves structural consistency, while Focal loss emphasizes hard pixels and improves small-object segmentation.

For dependency prediction, since only part of the categories contain valid dependency regions, we apply a focal-dice loss $\mathcal{L}_{\text{dep}}^{\text{fg}}$ on annotated samples and a cross-entropy loss~\cite{mao2023cross} $\mathcal{L}_{\text{dep}}^{\text{bg}}$ for the background:
\begin{eqnarray}
\mathcal{L}^{\text{IoU}}_{\text{dep}} = \mathcal{L}_{\text{dep}}^{\text{fg}} + \mathcal{L}_{\text{dep}}^{\text{bg}}.
\end{eqnarray}

An auxiliary loss $\mathcal{L}^{\text{aux}}_{\text{dep}}$ is similarly applied, and the total dependency loss is obtained by a weighted sum.

\section{Experiments}

\label{sec:expirement}

\subsection{Implementation Details}
\label{sec:expirement0}

We train and evaluate our model on the proposed AVAGD dataset using NVIDIA RTX 4090 GPUs. Audio features are extracted with VGGish \cite{hershey2017cnn} pretrained on AudioSet \cite{gemmeke2017audio}, and visual features with PVT-v2 \cite{wang2021pvtv2} pretrained on ImageNet-1K \cite{deng2009imagenet}. Both modalities are projected to 256 dimensions. Input images and masks are resized to $512\times512$. The model is trained for 25 epochs on 418K image–audio pairs using AdamW \cite{loshchilov2017decoupled} with a learning rate of $2\times10^{-5}$. To enhance generalization, we apply spatial (rotation, flipping) and color (brightness, contrast, saturation, hue) augmentations.  

For a fair comparison, AVS baselines originally designed for the S4 setting are adapted to our task. Audio clips are padded or cropped to 5 seconds, and a single visual frame is repeated five times. During evaluation, metrics are computed for each frame and averaged across the five frames.

\subsection{Comparison with SOTA Methods}
\label{sec:expirement2}
\begin{table*}[h]
\centering
\caption{Quantitative comparison with representative models from related methods on the AVAGD dataset. Evaluation metrics are mIoU and F1-score, where higher values are better. \textbf{Bold} indicates the best scores, while \underline{underline} denotes the second-best scores.}
\resizebox{2\columnwidth}{!}{%
\begin{tabular}{l|cccc|cccc|c}
\toprule
\multirow{2}{*}{Method}   & \multicolumn{4}{c|}{Seen} & \multicolumn{4}{c|}{Unseen} & \multirow{2}{*}{Reference}\\
 &$\text{mIoU}_{f}$ $\uparrow$ & $\text{F}_{f}$ $\uparrow$ & $\text{mIoU}_{d}$ $\uparrow$ & $\text{F}_{d}$ $\uparrow$ & $\text{mIoU}_{f}$ $\uparrow$ & $\text{F}_{f}$ $\uparrow$ & $\text{mIoU}_{d}$ $\uparrow$ & $\text{F}_{d}$ $\uparrow$ &\\
\midrule

TPAVI\cite{zhou2022audio} & 52.63&65.28 & 64.78&\underline{78.00} & 16.15 & 26.37& 47.00 & 68.71 & ECCV'22 \\
AVSegFormer\cite{gao2024avsegformer} &\underline{58.95} & \underline{72.48} & 65.84 & 76.35&15.22 & 25.96& 50.61 & 69.87 & AAAI'24\\
Selm\cite{li2024selm} & 55.95& 68.15& \underline{67.87} & 77.34& 18.39 & 27.88 & 50.79 &  71.25 & ACM MM'24\\
COMBO-AVS\cite{yang2023cooperation} &57.36 &72.15 & 65.70& 77.24&18.36 &27.30 & 50.56 & 72.63 & CVPR'24\\
Crab-7B\cite{du2025crab}  &58.89 &71.38 & 66.72& 76.97&\textbf{23.79} &\textbf{32.45} & \textbf{53.43} & \textbf{75.74} & CVPR'25\\
\textbf{AVAGFormer (Ours)}  & \textbf{63.08}&\textbf{74.56} & \textbf{69.52}& \textbf{80.27}&\underline{19.59} &\underline{29.74} & \underline{52.06} & \underline{74.09} & - \\
\bottomrule
\end{tabular}
}
\label{tab:tab2}
\end{table*}
We compare AVAGFormer with several SOTA methods adapted from AVS task, using the same PVT-v2 backbone and training settings for fair comparison. As shown in \tabref{tab:tab2}, AVAGFormer achieves the best performance across all metrics on both seen and unseen subsets. Compared to AVSegFormer~\cite{gao2024avsegformer} and COMBO-AVS~\cite{yang2023cooperation}, our model improves $\text{mIoU}_{f}$ by 4.13 and 5.72 respectively on the seen subset, and also shows clear gains in $\text{F-score}_{f}$, $\text{mIoU}_{d}$ and $\text{F-score}_{d}$. On the unseen subset, AVAGFormer maintains a consistent lead, highlighting its stronger generalization ability.

\subsection{Ablation Study}
\label{sec:expirement3}
\begin{table}[t]
\centering
\caption{\textbf{Effect of cross-modal siamese fusion.} V2A represents vision-to-audio fusion, and A2V represents audio-to-vision fusion.}
\resizebox{\columnwidth}{!}{%
\begin{tabular}{cc|cccc|cccc}
\toprule
\multirow{2}{*}{V2A} & \multirow{2}{*}{A2V} & \multicolumn{4}{c|}{Seen} & \multicolumn{4}{c}{Unseen}  \\
 &  &  $\text{mIoU}_{f}$ $\uparrow$ & $\text{F}_{f}$ $\uparrow$&  $\text{mIoU}_{d}$ $\uparrow$ & $\text{F}_{d}$ $\uparrow$ &  $\text{mIoU}_{f}$ $\uparrow$ & $\text{F}_{f}$ $\uparrow$&  $\text{mIoU}_{d}$ $\uparrow$ & $\text{F}_{d}$ $\uparrow$\\
\midrule
&  & 60.62 & 71.98&67.83& 78.92& 17.23 & 26.78&50.21&72.33\\
\ding{51} &  & \underline{62.48} & \underline{73.95}&\underline{68.92}& \underline{79.66}& \underline{19.03} & \underline{29.20}&\underline{51.17}&\underline{73.18}\\
 & \ding{51} & 61.93 & 72.76&68.21& 79.12& 17.95 & 27.46&50.63&72.74\\
\ding{51} & \ding{51} & \textbf{63.08} & \textbf{74.56}&\textbf{69.52}&\textbf{80.27} & \textbf{19.59} & \textbf{29.74}&\textbf{52.06}&\textbf{74.09} \\
\bottomrule
\end{tabular}
}

\label{tab:tab3}
\end{table}

\noindent\textbf{Effect of semantic-conditioned cross-modal mixer.} As shown in \tabref{tab:tab3}, both visual-to-audio (V2A) and audio-to-visual (A2V) fusion improve performance, validating the effectiveness of cross-modal interaction. V2A fusion moderately enhances both seen and unseen results by refining audio semantics, while A2V fusion yields larger gains, especially for unseen data, as audio-guided visual features better capture sound-related regions. Combining both directions achieves the best mIoU and F-scores, confirming that bidirectional fusion reinforces complementary features and strengthens cross-modal grounding and generalization.

\begin{table}[t]
\centering
\caption{\textbf{Comparison between CRA and CHA.} CRA represents cross-attention mixer, and CHA represents channel-attention mixer.}
\resizebox{\columnwidth}{!}{%
\begin{tabular}{c|cccc|cccc}
\toprule
\multirow{2}{*}{Attn} & \multicolumn{4}{c|}{Seen} & \multicolumn{4}{c}{Unseen}  \\
 &   $\text{mIoU}_{f}$ $\uparrow$ & $\text{F}_{f}$ $\uparrow$&  $\text{mIoU}_{d}$ $\uparrow$ & $\text{F}_{d}$ $\uparrow$ &  $\text{mIoU}_{f}$ $\uparrow$ & $\text{F}_{f}$ $\uparrow$&  $\text{mIoU}_{d}$ $\uparrow$ & $\text{F}_{d}$ $\uparrow$\\
\midrule
CHA&  62.40 & 73.87 & 68.34 & 79.83 & 18.74 & 29.01 & 51.30 & 73.80 \\
CRA& \textbf{63.08} & \textbf{74.56}&\textbf{69.52}&\textbf{80.27} & \textbf{19.59} & \textbf{29.74}&\textbf{52.06}&\textbf{74.09} \\
\bottomrule
\end{tabular}
}
\label{tab:tab_channel}
\end{table}

\noindent\textbf{Comparison between CRA and CHA.}
As shown in \tabref{tab:tab_channel}, both the cross-attention mixer (CRA) and channel-attention mixer (CHA) perform comparably, but CRA consistently achieves slightly higher results across metrics. This advantage stems from its explicit modeling of inter-modal dependencies, enabling finer alignment and more accurate localization of sound-related regions, while CHA mainly enhances intra-modal representation.

\begin{table}[t]
\centering
\caption{\textbf{Effect of dual-head affordance decoder.} SE represents squeeze-and-excitation, and MCA represents mask-conditioned attention.}
\resizebox{\columnwidth}{!}{%
\begin{tabular}{cc|cccc|cccc}
\toprule
\multirow{2}{*}{SE} & \multirow{2}{*}{MCA} & \multicolumn{4}{c|}{Seen} & \multicolumn{4}{c}{Unseen}  \\
 &&  $\text{mIoU}_{f}$ $\uparrow$ & $\text{F}_{f}$ $\uparrow$& $\text{mIoU}_{d}$ $\uparrow$& $\text{mIoU}_{d}$ $\uparrow$&  $\text{mIoU}_{f}$ $\uparrow$ & $\text{F}_{f}$ $\uparrow$& $\text{mIoU}_{d}$ $\uparrow$& $\text{mIoU}_{d}$ $\uparrow$\\
\midrule
&&  60.27 & 71.55 & 65.00 & 76.89& 16.80 & 25.98 & 49.88 & 71.95\\
\ding{51} && \underline{62.99} & \textbf{74.57} &  65.99 & 78.01  & \underline{19.12} & \underline{29.05} & 50.45 & 72.40\\
&\ding{51} & 60.37 & 71.69 &\underline{68.78} & \underline{79.43}&17.05 & 26.54 & \underline{51.72} & \underline{73.55}\\
\ding{51} &\ding{51} & \textbf{63.08} & \underline{74.56}&\textbf{69.52}&\textbf{80.27} & \textbf{19.59} & \textbf{29.74}&\textbf{52.06}&\textbf{74.09}\\
\bottomrule
\end{tabular}
}
\label{tab:tab4}
\end{table}

\noindent\textbf{Analysis on the design of affordance decoder.}
As shown in \tabref{tab:tab4}, removing either the SE block or the mask-conditioned attention (MCA) leads to a clear performance drop, indicating their complementary roles. The SE block enhances channel-wise feature selection, improving both function and dependency predictions, while MCA leverages the function mask to guide the dependency branch, facilitating recognition of relationship-dependent regions. Combining SE and MCA yields the best overall results and consistent gains on both seen and unseen subsets, demonstrating their effectiveness and efficiency with minimal computational overhead.
\begin{table}[t]
\centering
\caption{\textbf{Mutual enhancement between the two segmentation branches.} Dual indicates whether a dual-branch structure is used. Func and Dep denote supervision applied only to the function or dependency branch, respectively.}
\resizebox{\columnwidth}{!}{%
\begin{tabular}{ccc|cccc|cccc}
\toprule
\multirow{2}{*}{Dual} & \multirow{2}{*}{Func} & \multirow{2}{*}{Dep} & \multicolumn{4}{c|}{Seen} & \multicolumn{4}{c}{Unseen}  \\
 &&&  $\text{mIoU}_{f}$ $\uparrow$ & $\text{F}_{f}$ $\uparrow$& $\text{mIoU}_{d}$ $\uparrow$& $\text{mIoU}_{d}$ $\uparrow$&  $\text{mIoU}_{f}$ $\uparrow$ & $\text{F}_{f}$ $\uparrow$& $\text{mIoU}_{d}$ $\uparrow$& $\text{mIoU}_{d}$ $\uparrow$\\
\midrule
&\ding{51} &\ding{51}& 59.21 & 72.59 & 66.23 & 78.12&18.34&27.31&50.99&71.48 \\
\ding{51}&\ding{51} && \underline{61.65} & \underline{73.48} & 19.15 & 30.08& \underline{19.50} & \underline{29.60} & 18.20 & 33.10 \\
\ding{51}&&\ding{51} & 7.42 & 13.74 & \underline{68.86} & \underline{78.32}& 6.90 & 12.82 & \underline{51.08} & \underline{73.60} \\
\ding{51}&\ding{51} &\ding{51} & \textbf{63.08} & \textbf{74.56}&\textbf{69.52}&\textbf{80.27} & \textbf{19.59} & \textbf{29.74}&\textbf{52.06}&\textbf{74.09}\\
\bottomrule
\end{tabular}
}

\label{tab:mut_enh}
\end{table}

\noindent\textbf{Effect of dual-head decoding structure.}
As shown in \tabref{tab:mut_enh}, the proposed dual-head design fundamentally reshapes the decoder’s feature learning. Without it, the two tasks are learned independently, limiting shared semantic understanding. Introducing the dual-head structure encourages implicit cross-task priors, even under single-branch supervision, though improvements remain asymmetric. Joint optimization, however, enables synergistic interaction: the function head provides semantic priors that guide dependency localization, while the dependency head refines spatial and contextual consistency. This bidirectional feedback fosters structured cross-task supervision, leading to more generalizable representations and stronger transferability to unseen categories.

\begin{table}[t]
\centering
\caption{Effect of the weighting coefficient between $\mathcal{L}_{\text{aux}}$ and $\mathcal{L}_{\text{IoU}}$.}
\resizebox{\columnwidth}{!}{%
\begin{tabular}{c|cccc|cccc}
\toprule
\multirow{2}{*}{$\lambda$} & \multicolumn{4}{c|}{Seen} & \multicolumn{4}{c}{Unseen}  \\
 &  $\text{mIoU}_{f}$ $\uparrow$ & $\text{F}_{f}$ $\uparrow$& $\text{mIoU}_{d}$ $\uparrow$& $\text{F}_{d}$ $\uparrow$&  $\text{mIoU}_{f}$ $\uparrow$ & $\text{F}_{f}$ $\uparrow$& $\text{mIoU}_{d}$ $\uparrow$& $\text{F}_{d}$ $\uparrow$\\
\midrule
0   & 61.82 & 72.65 & 68.02 & 79.34 & 18.92 & 28.90 & 51.28 & 73.32 \\
0.1 & \textbf{63.08} & \underline{74.56}&\textbf{69.52}&\textbf{80.27} & \textbf{19.59} & \underline{29.74}&\textbf{52.06}&\textbf{74.09} \\
0.5 & \underline{62.45} & \textbf{74.60} & \underline{68.90} & \underline{79.92} & \underline{19.41} & \textbf{29.76} & \underline{51.95} & \underline{73.84} \\
1.0 & 61.54 & 72.41 & 68.31 & 79.18 & 19.27 & 29.51 & 51.72 & 73.55 \\
\bottomrule
\end{tabular}
}
\label{tab:lambda}
\end{table}

\noindent\textbf{Effect of the auxiliary loss weighting coefficient \(\lambda\).}
As shown in \tabref{tab:lambda}, incorporating the auxiliary loss consistently improves segmentation performance, peaking at $\lambda=0.1$ where both seen and unseen subsets achieve the best mIoU. A moderate auxiliary weight enhances gradient flow and stabilizes training, improving mask consistency and boundary precision. When $\lambda$ increases beyond 0.1, gains diminish, indicating that excessive auxiliary supervision may hinder optimization. Overall, $\lambda=0.1$ provides a good balance between accuracy and generalization.

\subsection{Case Study}

\label{sec:expirement4}

\begin{figure}[h]
    \centering
    \includegraphics[width=\linewidth]{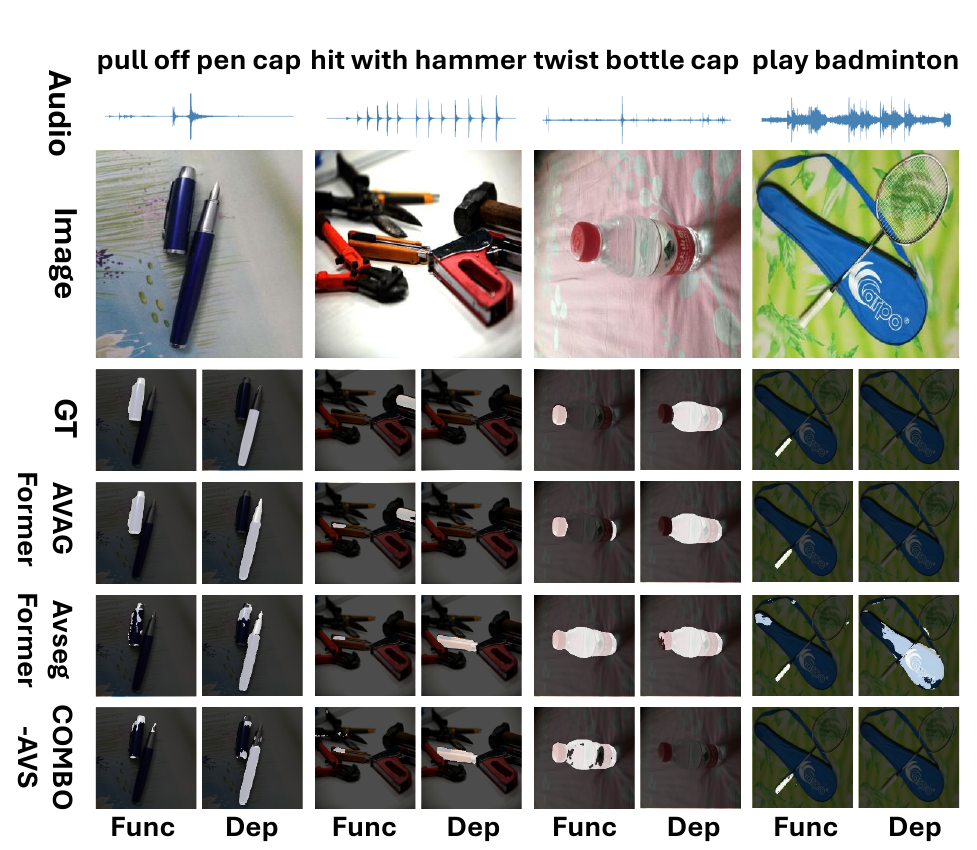}
    \caption{Qualitative results of AVAGFormer and other methods. The visual comparisons highlight that AVAGFormer produces more precise segmentation masks.}
    \label{fig:vis1}
\end{figure}
\begin{figure}[h]
    \centering
    \includegraphics[width=\linewidth]{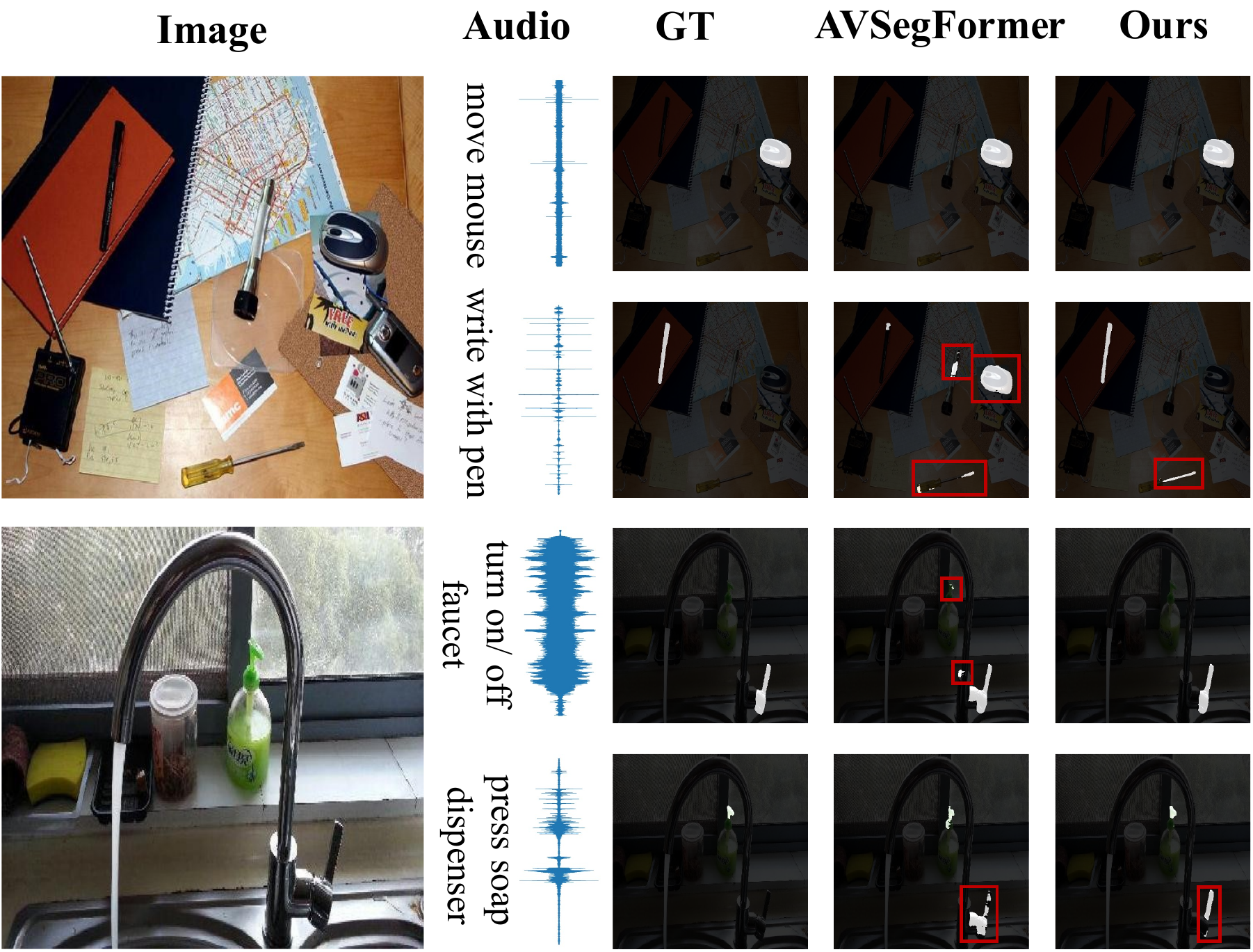}
    \caption{The comparison between AVAGFormer and AVSegFormer shows that when grounding affordance with different sounds on the same image.}
    \label{fig:vis2}
    \vspace{-3mm}
\end{figure}
As shown in \figref{fig:vis1}, AVAGFormer outperforms other methods in segmentation accuracy. It leverages cross-modal siamese fusion to accurately identify target objects, even in the presence of distracting objects, demonstrating a greater ability to focus on relevant features. This results in more precise segmentation in complex environments. Additionally, as shown in the challenging example in \figref{fig:vis2}, when grounding affordance based on different sounds within the same image, our model misidentifies fewer objects compared to AVSegFormer. This highlights the effectiveness and robustness of AVAGFormer in handling diverse, real-world scenarios.

\section{Conclusion}
\label{sec:conclusion}
In this paper, we introduce the Audio-Visual Affordance Grounding (AV-AG) task, accompanied by the AVAGD dataset, to advance research on multimodal affordance understanding. We also present AVAGFormer, which integrates cross-modal alignment and dual-head decoding to achieve state-of-the-art results. These findings demonstrate the potential of multimodal learning for fine-grained affordance reasoning. In future work, we plan to extend AVAGD to dynamic and multi-step interactions to better reflect real-world scenarios.
{
    \small
    \bibliographystyle{ieeenat_fullname}
    \bibliography{main}
}
\clearpage
\setcounter{page}{1}
\maketitlesupplementary

\section{Dataset}
\label{sec:appendix_section}
\subsection{Image Source}
Our images are primarily from ADE20K \cite{zhou2017scene,zhou2019semantic}, Open Images Dataset V7 \cite{kuznetsova2020open}, Products-10K \cite{bai2020products}, and Flickr \cite{flickr}. Additionally, to address the scarcity of images in the industry and agriculture domains, we also collect images from METU-ALET \cite{kurnaz2019alet}. The number of images from each source is shown in Table \ref{tab:sup_tab1}.

\begin{table}[h]
\centering

\begin{tabular}{ccc}
\toprule
Source & Number   & \%  \\
\midrule
  Products-10K & 3938 & 30.84 \\
  Open Images Dataset V7 & 3453 & 27.04 \\
  Flickr & 3268 & 25.60 \\
  ADE20K & 1305 & 10.22 \\
  METU-ALET & 804 & 6.30 \\
\bottomrule
\end{tabular}

\caption{
The number of images from each source.
}
\vspace{-0.05in}
\label{tab:sup_tab1}
\end{table}

\subsection{Dataset Division}
To evaluate the model's zero-shot generalization ability, we divide the entire dataset into two subsets: seen and unseen, based on affordance-object categories. The seen subset is further split into a training set (80\%) and a validation set (20\%), while all data in the unseen subset is used exclusively for testing. The affordance-object category splits are shown in Table \ref{tab:sup_tab2}.
\begin{table*}[t]
\centering
\begin{tabularx}{\textwidth}{lX}
\toprule
Subset & Affordance-Object Category  \\
\midrule
Seen & smooth@wood planner, snap@camera, shake@bottle, open or close@blinds, unlock@combination lock, pump@air pump, cut@sickle, play@string instruments, pull off@pen, open@key, dig@hoe, play@soccer, release or tighten@handbrake, press@soap dispenser, press@button, cut@knife, turn on or off@lighter, carve@knife, plunge@plunger, press@pen, open or close@drawer, ride@bike, pour@water, move@mouse, unfold@utility knife, chop@axe, cut@bolt cutter, open or close@door, play@flute, open or close@washing machine, turn on or off@gas stove, play@piano, prise@bottle opener, press@stapler, mop@mop, adjust@wrench, play@volleyball, stir@muddler, drill@electric drill, knock@door, clean@window, turn on@chainsaw, sharpen@pocket pencil sharpener, box@punching bag, spread@trowel, twist@knob, pull@zipper, hit@hammer, open@can, play@brass instruments, flush@toilet, clip@nail clipper, flip@bottle cap, lock@padlock, play@badminton, open or close@wave oven, brake@bike, push@window, twist@bottle cap, draw or write@pen, slide@door, twist@nut, press@mouse, play@drum, brush@toothbrush, sharpen@srank pencil sharpener, play@tennis, play@harp, cut@scissors, open or close@fridge, twist@screw, press@keyboard, brush@brush, spray@bottle, peel@peeler, turn on or off@faucet, play@guitar, brush@paint roller, cut@handsaw, pull@joystick, plug in or out@plug, turn on or off@switch, fasten@seatbelt, dig@shovel \\ 
Unseen & sweep@broom, cut@cutting machine, shift@gears, scrape@spatula, hit@meat hammer, brush@paint brush, play@basketball, play@guzheng, play@table tennis, kick@shuttlecock, play@erhu, twist@valve, play@accordion, twist@bottle opener, cut@chainsaw, open or close@car door, lock@door, slide@window, press@duster balloon, cut@garden scissors, press@garlic press, drill@jack hammer, shake@rattle drum, lock@slidelock, row@boat, scroll@mouse, ring@bike, flip@book, play@pipa \\
\bottomrule
\end{tabularx}

\caption{Affordance-object category splits, where affordance category is connected to object category with "@".}
\label{tab:sup_tab2}
\end{table*}

\subsection{More Statistics}
\begin{figure*}[h]
    \centering
    \includegraphics[width=0.9\linewidth]{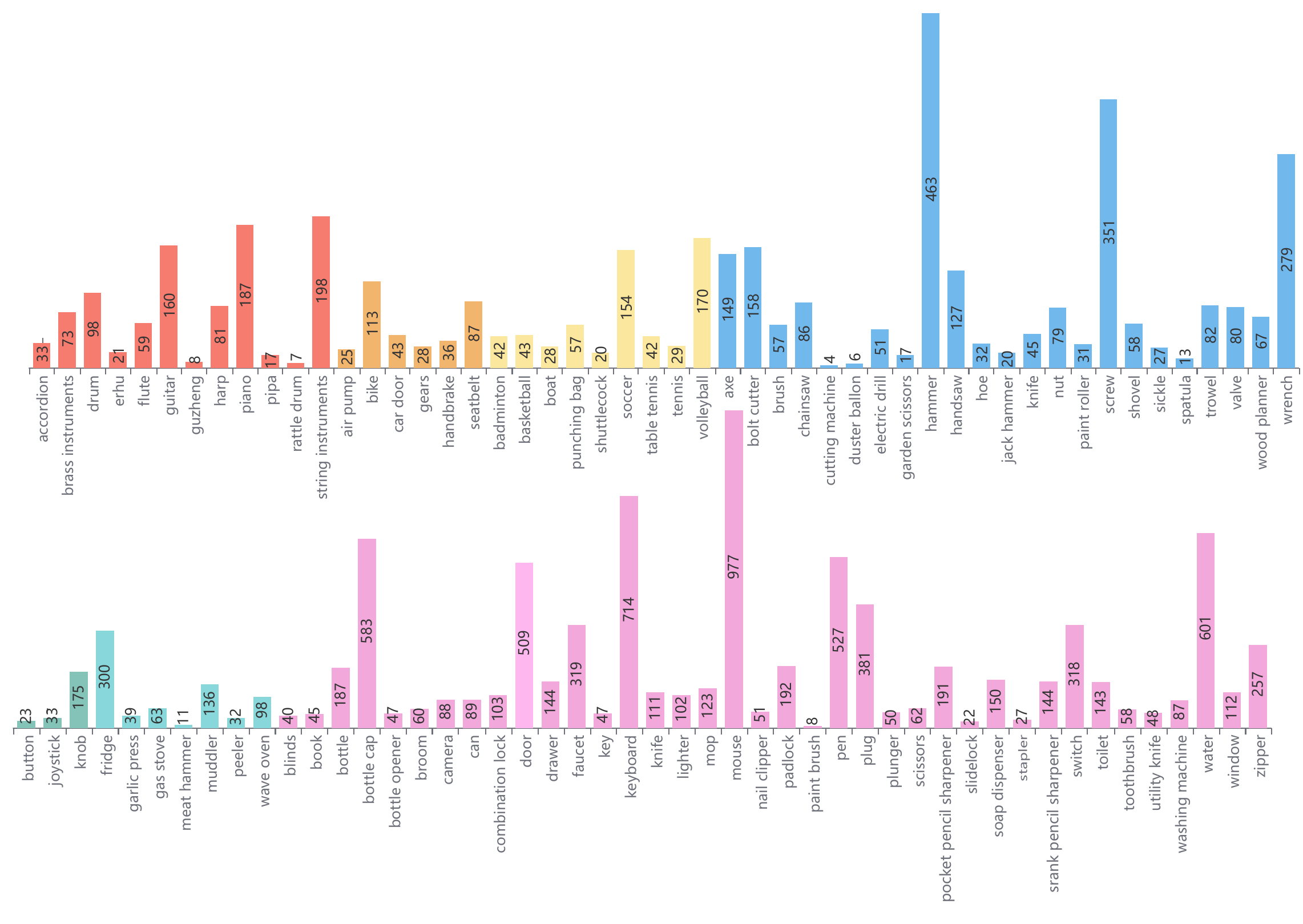}
    \caption*{(a) The number of images per object category in the AVAGD dataset.}
    \vspace{1em} 
    \includegraphics[width=0.9\linewidth]{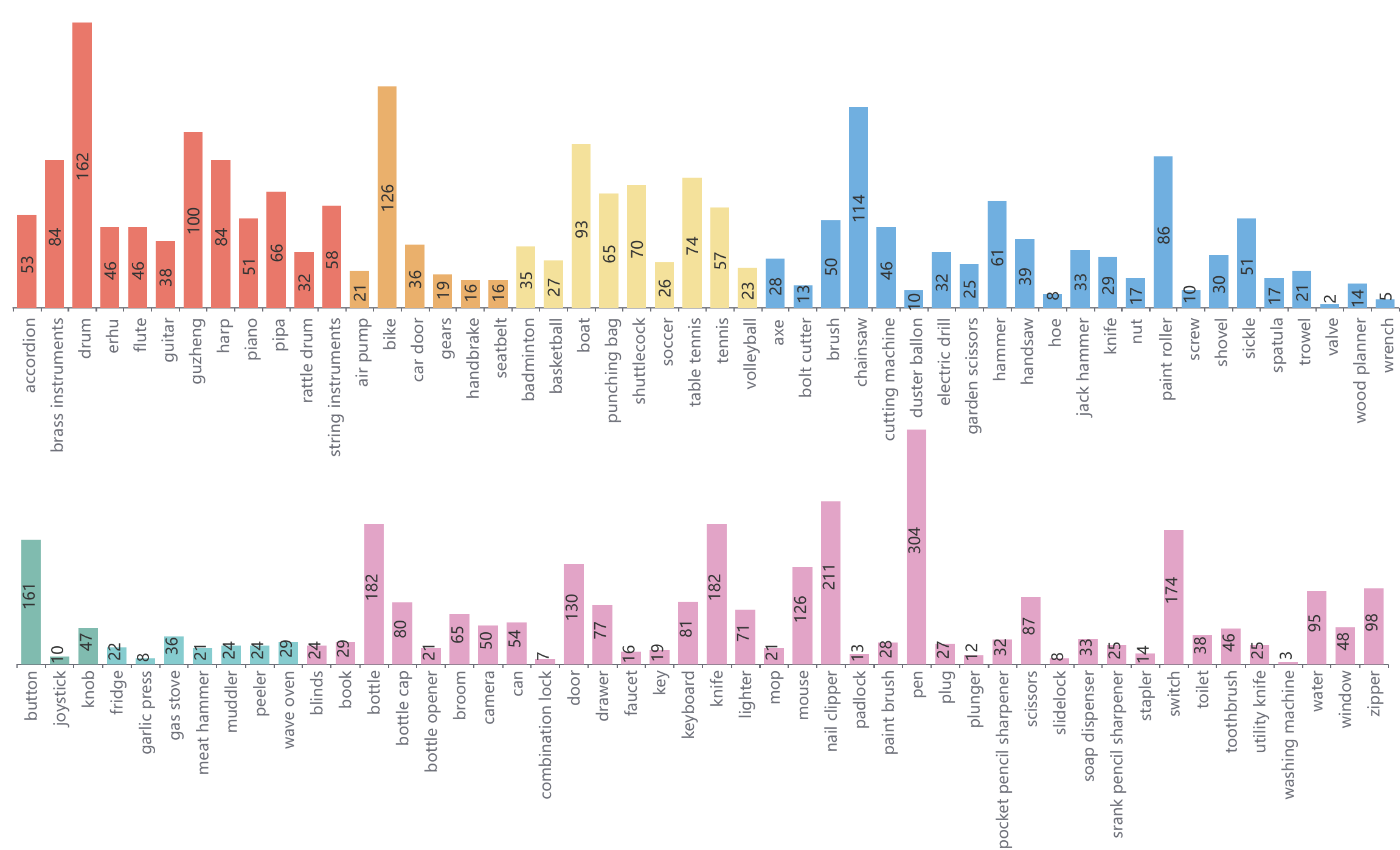}
    \caption*{(b) The number of audios per object category in the AVAGD dataset.}
    \caption{Statistics of the AVAGD dataset: (a) Number of images per object category. (b) Number of audios per object category.}
    \label{fig:sta_dataset}
\end{figure*}

\figref{fig:sta_dataset} provide a detailed overview of the number of images and audio clips for each object category.

\subsection{Evaluation Metrics}
In previous affordance grounding works, since the models only need to predict "action possibilities", they often use KLD, SIM, and NSS to measure the distributional correlation between the predicted heatmap and the ground truth. Our proposed task, however, sets higher expectations for the model, requiring it to perform pixel-wise segmentation of affordance regions. To evaluate the performance of our model, we adopt mIoU and F-Score as evaluation metrics, drawing from the AVS task.

\begin{itemize}
    \item \textbf{mIoU}: mIoU evaluates the pixel-wise overlap between the predicted segmentation mask and the ground truth mask. Given a predicted mask $P$ and a ground truth mask $Q$, IoU for each sample is calculated as:
    \begin{eqnarray}
    \text{IoU}(P, Q) = \frac{|P \cap Q|}{|P \cup Q|}
    \end{eqnarray}
    where $|P \cap Q|$ represents the number of overlapping pixels, and $|P \cup Q|$ is the total number of pixels in both masks. The mean IoU (mIoU) is then computed by averaging over all samples in the dataset:
    \begin{eqnarray}
    \text{mIoU} = \frac{1}{N} \sum_{i=1}^{N} \text{IoU}(P, Q)
    \end{eqnarray}
    where $N$ is the number of samples.

    \item \textbf{F-score}: F-Score is used to balance precision and recall for evaluating segmentation performance, particularly for unbalanced affordance regions. It is defined as:
    \begin{eqnarray}
    \text{F-score} = \frac{1}{N} \sum_{i=1}^{N} \frac{(1 + \beta^2) \cdot \text{Precision}_i \cdot \text{Recall}_i}{\beta^2 \cdot \text{Precision}_i + \text{Recall}_i}
    \end{eqnarray}
    where $\beta^2=0.3$ to emphasize recall. 
    
\end{itemize}

\section{Experiments}
\subsection{Comparison Methods}
\begin{itemize}
\item \textbf{TPAVI} \cite{luo2022learning}: TPAVI serves as the baseline for AVS. It addresses the non-co-occurrence of auditory and visual signals within the same frame by fusing these features across multiple frames using a structure similar to the non-local block. The fused features are then decoded by the Panoptic-FPN decoder.

\item \textbf{AVSegFormer} \cite{gao2024avsegformer}: Unlike TPAVI, which fuses multimodal features into a new feature, AVSegFormer employs a channel attention mixer to integrate audio features into visual features, enhancing the target representation ability of visual features. Additionally, inspired by Mask2Former, AVSegFormer adopts a dual decoder design with both pixel-level and query-level decoders, improving model performance and interpretability.

\item \textbf{Selm} \cite{li2024selm}: SelM introduces a selective mechanism using State Space models to suppress cross-modal noise and prevent audio-visual illusions. It features a dual alignment module for fine-grained early fusion and a cross-level decoder for layered reasoning, enabling more precise audio-guided segmentation and achieving state-of-the-art performance on AVS benchmarks.

\item \textbf{COMBO-AVS} \cite{yang2023cooperation}: COMBO-AVS incorporates SAM-generated masks as pixel-level priors to enhance visual feature modeling. It further introduces a Bilateral-Fusion Module to enable bidirectional alignment between audio and visual modalities, promoting more accurate cross-modal correspondence.

\end{itemize}
\subsection{Evaluation Pipeline}
\subsubsection{AVS Models}
In all our experiments, the models compared against AVAGFormer are adapted from those on the S4 task. Specifically, since the S4 task models take 5-second audio clips and 5 video frames as input, audio clips shorter than 5 seconds are repeated to reach the required length, while audio longer than 5 seconds is center-cropped to 5 seconds. For the visual input, a single frame is repeated five times. During evaluation, the scores of these five frames are averaged to obtain the final metric.
\subsection{Large Multimodal Models with Text Queries}
\begin{figure}[h]
    \centering
    \includegraphics[width=\linewidth]{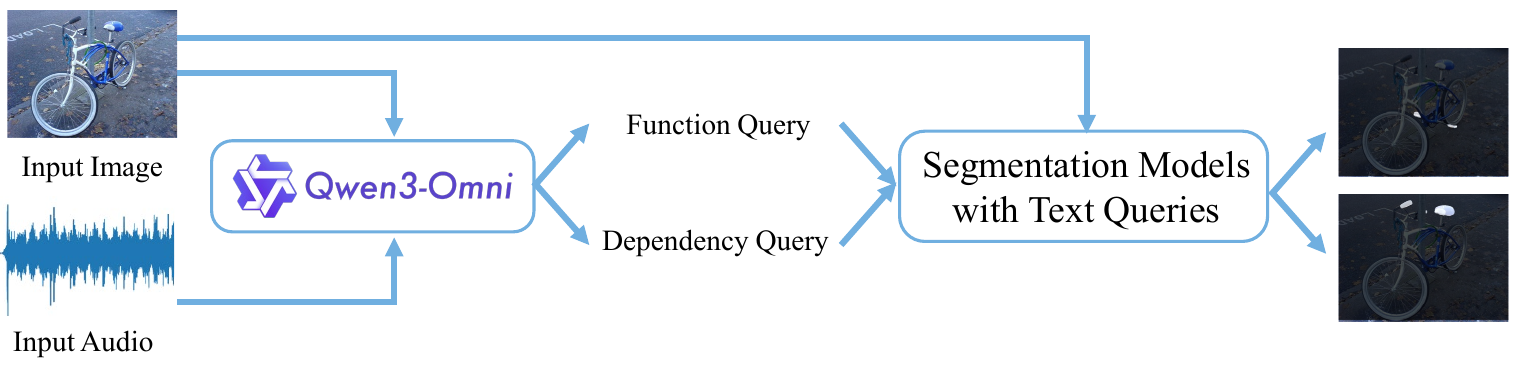}
    \caption{The evaluation pipeline for large multimodal segmentation models with text queries.}
    \label{fig: pp2}
\end{figure}
For large multimodal segmentation models with text queries, we design a two-stage evaluation protocol.
In the first stage, both the audio and the corresponding image are provided to Qwen3-Omni, which is instructed to generate textual descriptions for the function region and dependency region, respectively.
In the second stage, the generated textual descriptions are inserted into the segmentation queries of the corresponding models, serving as prompts to guide the final segmentation.
This design allows us to evaluate how effectively each model can leverage language-based affordance cues derived from multimodal inputs.
The prompt used for Qwen3-Omni is as follows:
\begin{lstlisting}
Assume you are a human user directing a robot to interact with an object. Given the provided image and audio, please identify:

- The functional region(s): the specific part(s) of the object that should be directly manipulated or operated to cause the action that leads to the sound you hear (note: this is the part being acted upon, not necessarily the sound-emitting part). 

- The dependency region(s): the contextual or supporting part(s) that indirectly enable or stabilize this affordance (e.g., structures that the action depends on), if any. 

For example: When hearing the sound of a bicycle being ridden, the functional region is the pedal (the part directly operated by the foot to produce the motion and thus the sound), while the dependency region is the handlebar (which supports balance and control but does not directly cause the sound). 

Output two concise robot instructions, with no additional explanation, in the following format: 

Instruction 1 (Functional): "Please find out [description of the functional region]." 
Instruction 2 (Dependency, if applicable): "Please find out [description of the dependency region]." 

\end{lstlisting}
\subsection{Large Multimodal Models with Audio Queries}
\begin{figure}[h]
    \centering
    \includegraphics[width=\linewidth]{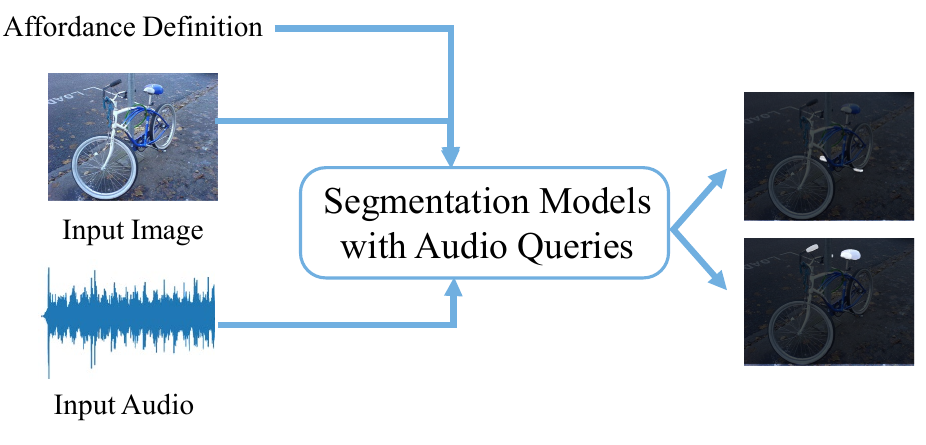}
    \caption{The evaluation pipeline for large multimodal segmentation models with audio queries.}
    \label{fig: pp1}
\end{figure}
For large multimodal segmentation models with audio queries, we adopt an end-to-end evaluation paradigm. Both the image and audio are directly fed into the model, while the system prompt explicitly provides the definitions of the function region and dependency region. The model is then instructed to perform segmentation for these two regions separately based on the given definitions and multimodal inputs. The system prompt we use is as follows:
\begin{lstlisting}
You are a helpful assistant guiding a robot to understand object affordances.

Given an image and an audio clip, identify and segment the following regions on the image:

- The functional region(s): the specific part(s) of the object that should be directly manipulated or operated to cause the action that leads to the sound you hear (note: this is the part being acted upon, not necessarily the sound-emitting part). 

- The dependency region(s): the contextual or supporting part(s) that indirectly enable or stabilize this affordance (e.g., structures that the action depends on), if any. 

Example:
When hearing the sound of a bicycle being ridden, the functional region is the pedal (the part directly operated by the foot to produce the motion and thus the sound), while the dependency region is the handlebar (which supports balance and control but does not directly cause the sound). 

\end{lstlisting}

\end{document}